\def\year{2020}\relax
\documentclass[letterpaper]{article} 
\usepackage{aaai20}  
\usepackage{times}  
\usepackage{helvet} 
\usepackage{courier}  
\usepackage[hyphens]{url}  
\usepackage{graphicx} 
\usepackage{graphicx}  
\newcommand{\citet}[1]{\citeauthor{#1} \shortcite{#1}} \newcommand{\citep}{\cite} 

\usepackage{latexsym}
\usepackage{caption}
\usepackage{subcaption}
\usepackage{amsmath}
\usepackage{multirow}
\usepackage{booktabs}
\usepackage{forest}
\usepackage{amssymb}
\forestset{
    nice empty nodes/.style={
        for tree={
            s sep=0.1em, 
            l sep=0.33em,
            inner ysep=0.4em, 
            inner xsep=0.05em,
            l=0,
            calign=midpoint,
            fit=tight,
            where n children=0{
               tier=word,
               minimum height=1.25em,
            }{},
            where n children=2{
               l-=1em,
            }{},
            parent anchor=south,
            child anchor=north,
            delay={if content={}{
                    inner sep=0pt,
                    edge path={\noexpand\path [\forestoption{edge}] 
                    			(!u.parent anchor) 
                               -- (.south)\forestoption{edge label};}
                }{}}
        },
    },
}

\usepackage[]{todonotes}  

\newcommand{\cumax}{\mathrm{cumax}}

\title{Inducing Constituency Trees through Neural Machine Translation}
\author{Phu Mon Htut$^1$\\
  {\tt pmh330@nyu.edu} \\\And
  Kyunghyun Cho$^{1,2}$ \\
  CIFAR Global Scholar \\
  {\tt kyunghyun.cho@nyu.edu} \\\And
  Samuel R. Bowman$^{1,2,3}$ \\
  {\tt bowman@nyu.edu} \\
  \AND
$^{1}$\normalfont Center for Data Science\\New York University\\60 Fifth Avenue\\New York, NY 10011\And
$^{2}$\normalfont Dept. of Computer Science\\New York University\\60 Fifth Avenue\\New York, NY 10011\And
$^{3}$\normalfont Dept. of Linguistics\\New York University\\10 Washington Place\\New York, NY 10003
  }

\begin{document}
\maketitle
\begin{abstract}

\textit{Latent tree learning} (LTL) methods learn to parse sentences using only indirect supervision from a downstream task. Recent advances in latent tree learning have made it possible to recover moderately high-quality tree structures by training with language modeling or auto-encoding objectives. 
In this work, we explore the hypothesis that decoding in machine translation, as a conditional language modeling task, will produce better tree structures since it offers a similar training signal as language modeling, but with more semantic signal.
We adapt two existing latent-tree language models---PRPN and ON-LSTM---for use in translation. We find that they indeed recover trees that are better in F1 score than those seen in language modeling on WSJ test set, while maintaining strong translation quality. We observe that translation is a better objective than language modeling for inducing trees, marking the first success at latent tree learning using a machine translation objective. Additionally, our findings suggest that, although translation provides better signal for inducing trees than language modeling, translation models can perform well without exploiting the latent tree structure.   

\end{abstract}

\section{Introduction} 
\label{sec:intro}

The distribution of words of natural language sentences exhibits an implicit hierarchical structure \citep{Chomsky1957SyntacticS}. 
\textit{Grammar induction}, the task of discovering this latent syntactic structure of language without explicit supervision, is a long-standing open problem in computational linguistics and natural language processing \citep{Klein2001,Klein2002,Smith2005}. Successes in this area could potentially inform linguistic work on grammar acquisition and provide the evidence on the question of \textit{the poverty of the stimulus}  \citep{Chomsky1957SyntacticS,Clark01CFG}. 

Incorporating syntactic parse trees, the objects that grammar induction systems aim to discover, offers some benefits to neural models for NLP tasks such as semantic role labeling \citep{Strubell18LISA}, and machine translation \citep{Aharoni2017TowardsSN}. These models obtain this parse information from parsers trained on separate treebanks. However, since human annotation is expensive, these treebanks are generally small and offer limited coverage of domains and languages. This motivates the work in \textit{latent tree learning} (LTL), a variant of grammar induction that trains a parser as part of a larger neural network model. 

Recent work using a language modeling objective represents the first real success with LTL. The Parsing-Reading-Predict Network (PRPN; \citet{shen2018neural}) uses a modified self-attention  mechanism guided by a convolution-based parser to incorporate latent syntax in an LSTM-based language model. The Ordered Neurons model (ON-LSTM; \citet{shen2018ordered}), proposed in follow-up work, adds a different kind of gating-based inductive bias that enforces a hierarchical order to the hidden state neurons of an LSTM unit and substantially outperforms PRPN in terms of F1 score on the test set of the Wall Street Journal section of Penn Treebank (WSJ; \citet{DBLP:journals/coling/MarcusSM94}) when trained as a language model. 

Other attempts at LTL that use supervised semantics-oriented natural language inference have succeeded at producing effective task models, but have not produced effective parsers \citep{YogatamaBDGL16,choi2017unsupervised,Williams2018a}. As these models focus mainly on classification, the classification signal alone might not be sufficient for inducing constituency trees. In concurrent work, \citet{LiMK19} achieve large gains in F1 by supplementing the Tree-RNN NLI classification model of \citet{choi2017unsupervised} with parser from PRPN language model using imitation learning, outperforming the parsing F1 of both Tree-RNN and PRPN on sentences from natural language inference datasets. This prompts two questions: Which other objectives have viable signals for LTL? Is it possible to perform better than plain language models by using additional, potentially more semantically informed, training signal from a labeled-data task to guide LTL?

Since we expect machine translation to require some understanding of syntax and semantics, we hypothesize that it could provide sufficient signal for LTL. In this work, we adapt the two existing high-performing LTL language models---PRPN and ON-LSTM---as decoders for German--English (De-En) and Chinese--English (Zh-En) translation. We investigate (i) whether machine translation is a good downstream task for LTL, and (ii) whether latent tree information improves translation quality. We observe that both PRPN and ON-LSTM decoders perform better or comparably to the baseline LSTM decoder in translation BLEU score. 

Our analysis shows that the trees generated by the NMT decoders are reasonably similar (measured in terms of F1 score) to the established Penn Treebank formalism. However, even though the variance across restarts in BLEU is low, we find that the variance in parsing F1 is extremely high, suggesting that these models are not robust. Despite this high variance in F1, both PRPN and ON-LSTM trained as decoders for NMT  produce better or comparable F1 scores across restarts than when they are trained as language models.  Thus, we conclude that machine translation offers a richer supervision signal for inducing constituency trees than language modeling, and represents the first success for LTL with a machine translation objective.

\section{Related Work} 
\label{sec:related}
The previous work on LTL with labeled data tasks includes classification models \citep{YogatamaBDGL16,maillard2017jointly,choi2017unsupervised} that are designed as sentence embedding models, whose composition order is guided by a latent tree structure, trained on natural language inference (SNLI/MNLI; \citet{snli:emnlp2015},\citet{WilliamsNB17}) or the Stanford sentiment treebank (SST; \citet{SocherPWCMNP13}). \citet{Williams2018a} reports that these models fail to learn linguistically plausible constituency trees. More recently, latent tree models that use language modeling \citep{shen2018neural,shen2018ordered} or auto-encoding \citep{Drozdov2018DIORA,Kim2019URNNG,KimCompound19} have achieved encouraging results. 

Our work is also related to work on machine translation models that use, or jointly learn, parse trees. Previous work has attempted to incorporate syntactic information in machine translation using annotated gold parses \citep{Eriguchi2017LearningTP,Bastings2017GraphCE}. Additionally, there are attempts to incorporate latent parse information in NMT models. However, all these models fail to induce consistent, non-trivial trees. \citet{Bradbury2017TowardsNM} propose an RNNG-based \citep{DyerRNNG16} encoder and decoder trained with REINFORCE to induce trees through NMT. 
\citet{Tran2018InducingGW} propose an NMT model that uses a self-attention encoder to induce latent dependency trees on the source side and report that the induced trees are task-specific and do not conform to the conventional definition of syntax. \citet{Bastings2019ModelingLS} investigate the conditions where induced latent tree structures can benefit NMT by adding a latent dependency parser-like graph component to CNN and RNN based NMT models. Although they achieve substantial gains in BLEU, they report that the induced trees are largely trivial.

\section{Models}
\label{sec:methods}

We use a bidirectional LSTM as the encoder in all our experiments. We provide our adaptation of PRPN and ON-LSTM as the latent-tree based NMT decoders in this section. 
We refer readers to the original papers for the two LTL methods for a complete motivation for each model architecture, but we recap both architectures here, with a focus on our MT-specific modifications.

\subsection{Parsing-Reading-Predict Network (PRPN) Decoder} 
\label{sec:PRPNdecoders}

PRPN is made up of three components. We keep the first component (the parser), and made modifications to the other two components, which we will describe in this section.
PRPN consists of a \textit{convolution-based parser} that measures how syntactically related two consecutive pairs of tokens in a sentence are, and uses this syntactic distance to divide a sentence into constituents. 
The syntactic distance $d_i$ between two consecutive pairs of word embeddings $e_{i-1}$ and $e_{i}$ is computed by running a convolutional kernel over a set of $L$ previous tokens and the current token ($e_{i-L}$, $e_{i-L+1}$,..., $e_{i}$).  The kernel size $L$ represents a look-back range, the amount of immediate history that the parser can take into account when calculating the syntactic distance $d_i$. 
Mathematically, syntactic distance $d_{i}$ between $e_{i-1}$ and $e_{i}$ is computed as: 
\begin{equation*}
\label{conv_kernel_1}
h_i^{parser} = \mathrm{ReLU}(W_c \left[ \begin{matrix} e_{i-L} \\ e_{i-L+1} \\ ... \\ e_i \end{matrix} \right] + b_c)
\end{equation*}
\begin{equation*} \label{conv_kernel_2}
d_i = \mathrm{ReLU} \left(W_d h_i^{parser} + b_d\right)
\end{equation*}
where $W_c$ and $b_c$ are the kernel parameters. $W_d$ and $b_d$ can be seen as another convolutional kernel with window size 1, convolved over $h_i^{parser}$.

PRPN determines the closest word $y_j$ that has larger syntactic relationship than $d_j$ for time step $t$ by computing $\alpha_{j}^{t}$:
\begin{equation*}  \label{soft_alpha}
\alpha_j^t = \frac{\mathrm{hardtanh} \left( (d_t - d_{j}) \cdot \tau \right) + 1}{2} 
\end{equation*}
where $\tau$ is the temperature parameter that controls the sensitivity of $\alpha_j^t$ to the differences between distances. The soft gate values that will be used for language modeling are then computed as:
\begin{equation*} \label{gate_equal_multialpha}
g_{i}^{t} = \mathbf{P}(l_{t}\leq i) = \prod_{j=i+1}^{t-1}\alpha_j^{t}
\end{equation*}

The next component is a \textit{reading network} which is an RNN-based language model with a self-attention gating mechanism. PRPN uses an \textit{LSTM network} (LSTMN; \citet{Cheng2016LongSM}), which is a modified LSTM that replaces its memory cell with a memory network \citep{Weston2015MemoryN}, as the core component of the language model. At each time-step, the reading network links the current time-step with all the previous time-steps that are syntactically related using structured self-attention:
\begin{align*}
k_t &= W_h h_{t-1}^{dec} + W_x x_t \\
\tilde{s}_i^t &= \mathrm{softmax}(\frac{h_i^{dec} k_t^{\mathrm{T}}}{\sqrt{\delta_k}})
\end{align*}
where, $\delta_k$ is the dimension of the hidden state. The structured intra-attention weight is defined based on the gates $g_i^t$:
\begin{align*}
s_i^t &= \frac{g_i^t \tilde{s}_i^t}{\sum_i g_i^t}
\end{align*}
An adaptive summary vector for the previous hidden tape and memory of LSTMN denoted by $\tilde { h } _{ t }^{dec}$ and $\tilde { c } _{ t }^{dec}$ are computed as:
\begin{equation*}
\left[ \begin{matrix} \tilde { h } _{ t }^{dec} \\ \tilde { c } _{ t }^{dec} \end{matrix} \right] = \sum _{ i=1 }^{ t-1 } s_{ i }^{ t }\cdot m_i = \sum _{ i=1 }^{ t-1 } s_{ i }^{ t }\cdot \left[ \begin{matrix} h_i^{dec} \\ c_i^{dec} \end{matrix} \right] 
\end{equation*}

The reading network then takes $e_t$ (the current decoder input embedding), $\tilde { c } _{ t }^{dec}$ and $\tilde { h } _{ t }^{dec}$ as input, computes the values of $c_t^{dec}$ and $h_t^{dec}$ by the LSTM recurrent update \citep{Hochreiter1996}.
Then, the \textit{write} operation concatenates $h_t^{dec}$ and $c_t^{dec}$ to the end of hidden and memory tape. In our MT adaptation, the reading network of the decoder will take the previous hidden state, performs attention mechanism \citep{Bahdanau2015NeuralMT} on encoder hidden states to get the encoder context vector, and use the concatenation of previous decoder hidden state and encoder context vector to compute current hidden state.

In the original PRPN model, a \textit{prediction network} is used to estimate the syntactic distance between the current word and unobserved future word; PRPN considers this distance in calculating the language model probabilities to account for the syntactic relation between the current state and the unobserved future word. In our adaptation, we use a simple feedforward neural network, in place of prediction network, on current decoder hidden state to predict the next decoder output; we do not observe a significant loss in performance by doing this.

\subsection{Ordered Neurons (ON-LSTM) Decoder}
\label{sec:ONLSTMdecoders}

Unlike PRPN, which uses an additional convolution-based parser to guide the LSTM language model, the ON-LSTM incorporates a syntax-based inductive bias into the LSTM unit itself. The ON-LSTM assumes that the hidden state represents all nodes on the path between the current leaf node and the root node, and that the different nodes on the path are represented by the different chunks of adjacent neurons in the hidden state. The ON-LSTM is designed to dynamically allocate a different number of hidden state neurons to different nodes by using a master input gate and a master forget gate. 

For a binary gate $g=(0, ..., 0, 1, ...,1)$, the probability of the k-th value in g being 1 can be defined as:
\begin{align*}
    p(d) &= \mathrm{softmax}(\ldots) \\
    p(g_k=1) &= p(d \leq k) = \sum_{i \leq k} p(d = i)
\end{align*} where $d$ is the index of the first 1 in $g$. Based on this, the authors propose a \textit{cumax}, cumulative sum of softmax, function to define the gating mechanism that splits the hidden state into 0 and 1 segments. 
Using the $\cumax$ function, the master forget gate and the master input gate are defined as:
\begin{align*}
    \Tilde{f}_{t} &= \cumax (W_{\Tilde{f}}x_{t}+U_{\Tilde{f}}h_{t-1}+b_{\Tilde{f}}) \\ 
    \Tilde{i}_{t} &= 1 - \cumax (W_{\Tilde{i}}x_{t}+U_{\Tilde{i}}h_{t-1}+b_{\Tilde{i}}) 
\end{align*}


The master forget gate is responsible for the erasing behavior and  of the hidden state neurons , and the values in the master forget gate are restricted to monotonically increase from 0 to 1 (for example, [0, .., 0, 0.1, .., 0.9, 1]). The master input gate is responsible for the writing behavior of the hidden state neurons.  and the values of this gate are restricted to monotonically decrease from 1 to 0. 
A large fraction of 1's in the master input gate means the model is preserving long term information. A large fraction of 0's in the master forget gate means the ON-LSTM is erasing a large chunk of hidden neurons, which indicates the end of a high level constituent of the tree. Thus, each layer $L$ of ON-LSTM can induce a latent constituency tree by calculating the depth $\hat{d}_t^{f(L)}$ of a node $x_t$ based on the value of the master forget gate at time $t$: 
\begin{equation*}
    \hat{d}^{f(L)}_t = \mathbb{E} \left[ d^{f(L)}_t \right] 
\end{equation*}

In our ON-LSTM NMT decoder, we apply an attention mechanism using this hidden state and the encoder's hidden states to predict the translated output token. Following the original work, our implementation of ON-LSTM has 3 layers, and each layer of ON-LSTM produces a constituency tree. We report results for each layer separately.

\begin{table*}[t]
\small
\centering
\setlength{\tabcolsep}{4.2 pt} 
\begin{center}
\begin{tabular}{l|rr|rr|rr|r}
\toprule
 \multirow{ 3}{*}{\bf Model} &   \multicolumn{4}{c|}{\bf IWSLT'14 De-En: Translation } & \multicolumn{3}{c}{\bf IWSLT'14 De-En: Language Modeling} \\ 
    & \multicolumn{2}{c|}{ \bf BLEU } &  \multicolumn{2}{c|}{ \bf F1 (Target) } & \multicolumn{2}{c|}{ \bf F1 (Target)   } &  \bf Perplexity  \\
 & \bf Median$\:(\sigma)$ & \bf max & \bf Median$\:(\sigma)$ & \bf max & \bf Median$\:(\sigma)$ & \bf max & \bf Median$\:(\sigma)$  \\
\midrule
 \multicolumn{8}{c}{\bf Word models}  \\ 
 \midrule
 LSTM  &  26.1 (0.3) & 26.1 & -- &  -- & -- & -- & --  \\
 PRPN  & \bf 29.6 (0.5) & \bf 30.2 &  \bf 53.0 (15.2) & \bf 56.1 & \bf 44.6 (3.0) & \bf 48.4 & 74.1 (0.2)  \\
 ON-LSTM   & 28.3 (0.9) & 28.8 &  -- & -- & -- &  -- & \bf 68.8 (0.3)  \\
 \hspace{1.5em}Layer 1  & -- & --  & \underline{ 42.4 } (13.5)  & \underline{ 45.5 } & 20.4 (2.6) & 24.0 &  -- \\
 \hspace{1.5em}Layer 2  & -- &  --  & 28.5 (16.5) & 49.4 & \underline{ 38.1 } (1.3) & \underline{ 38.8 } & --  \\
 \hspace{1.5em}Layer 3 & -- & --  &  19.7 (12.5) & 45.3 & 27.4 (4.9) & 32.5 & -- \\
 \midrule
  \multicolumn{8}{c}{\bf BPE models}  \\ 
 \midrule
 LSTM  & 30.7 (7.8) & 31.1 & -- &  --  & -- & -- & --   \\
 PRPN  & \bf 31.0 (0.0) & \bf 31.4 &  \bf 47.0 (2.4) & \bf 51.0 & -- &  -- & --  \\
 ON-LSTM   & 29.9 (0.1) &  30.8   & -- & -- & -- &--  &  --  \\
 \hspace{1.5em}Layer 1  & -- & --  & \underline{36.7} (7.8) & \underline{42.7} & -- & -- & --  \\
 \hspace{1.5em}Layer 2  & -- & --   & 24.2 (10.1) & 35.4 & -- & -- & --  \\
 \hspace{1.5em}Layer 3 & -- &  --  & 28.7 (13.2) & 42.8 & -- & -- & --  \\
 \bottomrule 
\end{tabular}
\end{center}
\caption{\label{tab:result-table} Downstream task performance and parsing F1 for NMT models trained on IWSLT'14 De-En and language models trained on the English half of IWSLT'14 De-En. We compute F1 (Target) using the parses that are produced by feeding the source sentences and gold reference target sentences of the IWSLT'14 De-En test set into each NMT model. Bold marks the best performance for each column. Underlining marks the F1 scores produced by the best parsing layer of each ON-LSTM model. We report the median, standard deviation, and max scores across 5 restarts.
} 
\end{table*}

\begin{table}[t]
\small
\centering
\setlength{\tabcolsep}{4 pt} 
\begin{center}
\begin{tabular}{lrrrr}
\toprule
  \bf Model  & \multicolumn{2}{c}{ \bf BLEU } & \multicolumn{2}{c}{ \bf F1 (Target) } \\
  & \bf Median$\:(\sigma)$ & \bf max &  \bf Median$\:(\sigma)$ & \bf max \\
 \midrule
 \multicolumn{5}{c}{\bf Word models}  \\ 
 \midrule
 LSTM  & 15.1 (0.0) & 15.7 & -- & -- \\
 PRPN  & 14.8 (0.4) & 15.3 &  37.9 (12.5) &  54.2 \\
 ON-LSTM   & \bf 16.2 (0.3) & \bf 16.4  &  -- & -- \\
 \hspace{1.5em}Layer 1  & -- & -- & 23.1 (10.8) & 36.8 \\
 \hspace{1.5em}Layer 2  & -- & -- & \bf \underline{47.0} (13.6) & \bf \underline{54.9} \\
 \hspace{1.5em}Layer 3 & -- & -- & 15.0 (9.5) & 36.3 \\
 \midrule
  \multicolumn{5}{c}{\bf BPE models}  \\ 
 \midrule
 LSTM  & \bf 15.8 (0.0) & 15.9 &  -- & --  \\
 PRPN  &   15.5 (0.0) & \bf 16.2 &  46.7 (0.9) & 48.0  \\
 ON-LSTM  &  15.4 (0.3) & 15.7 & -- & -- \\
 \hspace{1.5em}Layer 1  & -- & -- &  \bf \underline{41.9} (11.9) & \bf \underline{49.9} \\
 \hspace{1.5em}Layer 2  & -- & -- &  29.2 (7.5) & 41.5 \\
 \hspace{1.5em}Layer 3  & -- &  --&  29.4 (11.2) & 40.3 \\

 \bottomrule 
\end{tabular}
\end{center}
\caption{\label{tab:result-table-zh} Machine translation performance (BLEU) and parsing F1 with respect to the Stanford parser on the IWSLT'17 Zh-En test set, following the convention used in Table \ref{tab:result-table}. ON-LSTM achieves the best parsing F1 on this dataset, whereas PRPN achieves the best parsing performance on IWSLT'14 De-En (Table \ref{tab:result-table})
}
\end{table}

\begin{figure*}[t!]
    \centering
    \begin{subfigure}[t]{0.5\textwidth}
        \centering
        \includegraphics[height=1.8in] {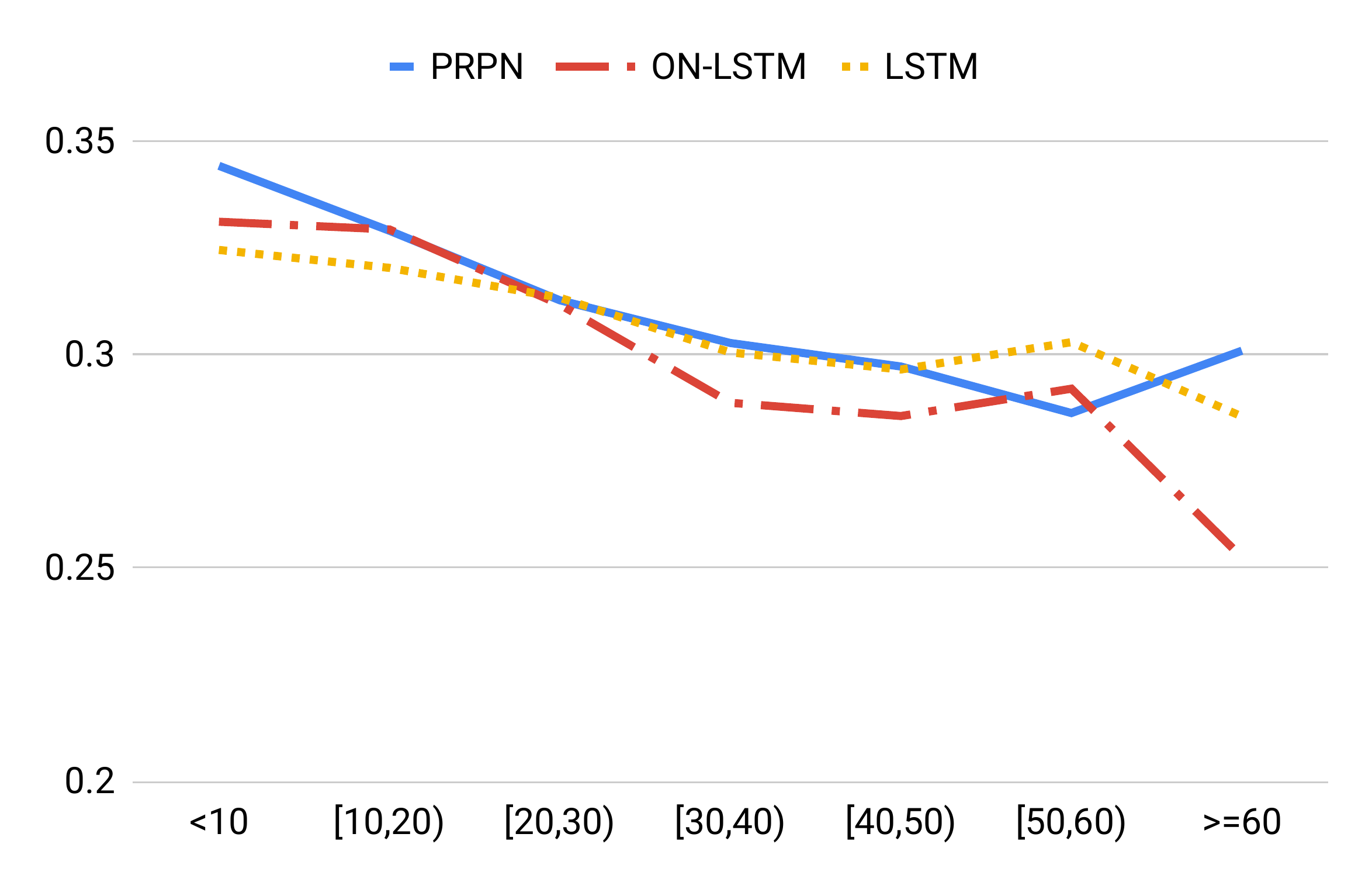}
        \caption{Models trained on IWSLT'14 De-En with BPE tokens. }
    \end{subfigure}%
    ~ 
    \begin{subfigure}[t]{0.5\textwidth}
        \centering
        \includegraphics[height=1.8in] {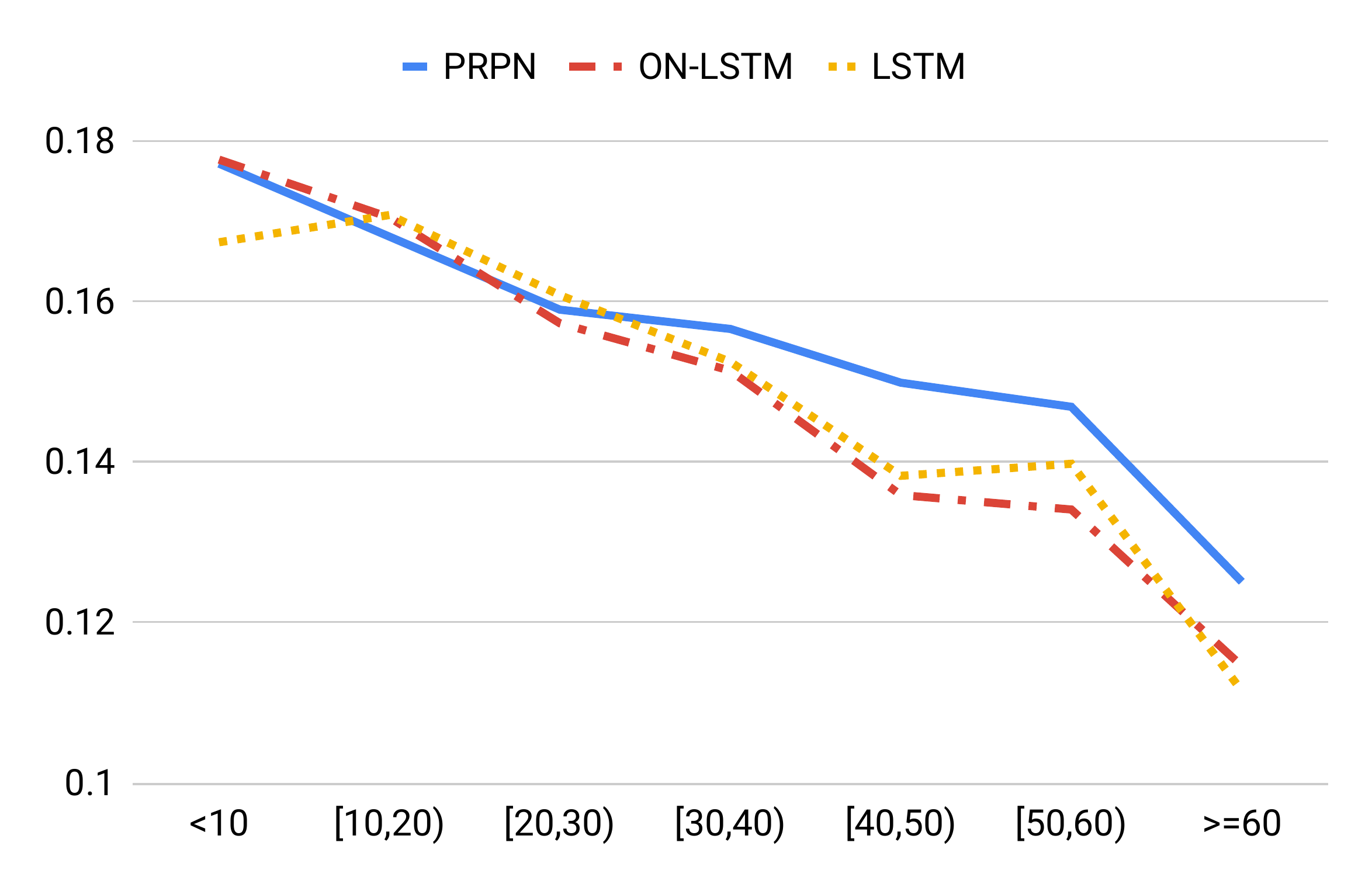}
        \caption{Models trained on IWSLT'17 Zh-En with BPE tokens.}
    \end{subfigure}
    \caption{Comparison of average sentence-level smoothed BLEU scores of the models for sentence length buckets. We choose the model with maximum BLEU score on each dataset for each model type trained with BPE tokens on both source and target sides of NMT. The horizontal axis represents the sentence length baskets and vertical axis represents the average sentence-level smoothed BLEU. For both datasets, PRPN performs competitively with LSTM across the different sentence length baskets. ON-LSTM performs poorly for long sentences.
    }
    \label{fig:bleu-length}
\end{figure*}

\begin{table}[t]
\small
\centering
\setlength{\tabcolsep}{2 pt} 
\begin{center}
\begin{tabular}{lrrrrr} 
\toprule
  \bf Model  & \bf Train &  \bf Perplexity  & \multicolumn{2}{c}{ \bf F1  }  \\
 & \bf Data &  \bf Median$\:(\sigma)$ & \bf Median$\:(\sigma)$ &  \bf max \\
 \midrule
 PRPN  & WSJ (LM) & 97.6 (0.6) & \bf 32.5 (6.7) &  \textbf{36.6}   \\
 ON-LSTM & WSJ (LM)   & \bf 81.1 (0.6) & --  &  --  \\
 \hspace{0.5em}Layer 1  & -- & -- & 21.9 (2.2)  & 26.0  \\
 \hspace{0.5em}Layer 2   & -- & -- &  20.5 (7.3) & 33.0  \\
 \hspace{0.5em}Layer 3  & -- & -- &  25.1 (6.8) & 36.1   \\
 \midrule
 PRPN  & IWSLT (LM) & 74.1 (0.2) & \bf 37.1 (8.5) & \textbf{46.8}   \\
 ON-LSTM & IWSLT (LM) & \bf 68.8 (0.3) &  -- &   --  \\
 \hspace{0.5em}Layer 1  & -- & -- & 19.1 (2.7)  & 21.8  \\
 \hspace{0.5em}Layer 2  & -- & -- & \underline{34.6} (1.9) & \underline{35.8}  \\
 \hspace{0.5em}Layer 3  & -- & -- &  21.4 (6.1) & 31.1   \\
 \midrule
 PRPN  & IWSLT (MT) & -- & \bf 43.7 (13.2) & \textbf{46.9}   \\
 ON-LSTM & IWSLT (MT) & -- &  -- &   --  \\
  \hspace{0.5em}Layer 1  & -- & -- & \underline{38.7} (12.6)  & \underline{46.2}  \\
  \hspace{0.5em}Layer 2  & -- & -- &  26.3 (15.6) & 49.2  \\
  \hspace{0.5em}Layer 3  & -- & -- & 19.8 (6.8) & 35.6   \\
  \midrule
 Random  & -- & -- & \bf 21.3 (0.0) & 21.4   \\
 Balanced & -- & -- & \bf 21.3 (0.0) & 21.3   \\
  \bottomrule  
\end{tabular}
\end{center}
\caption{\label{tab:lm-table} Language modeling perplexity and parsing F1 on the WSJ test set. We use the WSJ training set and the English sentences of IWSLT'14 De-En to train the language models. The translation models are trained on the IWSLT'14 De-En dataset. NMT use the \texttt{[EOS]} token as source input in NMT model and WSJ as target sentences. We report the median, standard deviation, and max scores across restarts. Bold marks the best performance for each column. Underlining marks the F1 scores produced by the best parsing layer of each ON-LSTM model. 
}
\end{table}

\section{Experimental Setup}

\subsection{Baselines}

We implement an LSTM decoder with attention \citep{Bahdanau2015NeuralMT} as our baseline translation model. To compare the parsing performance of NMT decoders with language models, we also train PRPN and ON-LSTM as language models. For language modeling experiments, we implement the smaller version of the PRPN, the PRPN-UP model used in \citet{Htut2018GrammarIW}, and  succeed in replicating the perplexity of 97.6 (Table \ref{tab:lm-table}) on the WSJ test set reported in their paper. However, we fail to replicate the 56.2 perplexity of the ON-LSTM language model reported in \citet{shen2018ordered} on WSJ, possibly due to a difference in hyperparameters. 

\subsection{Data and Preprocessing} 

We train the language models separately on the English sentences of the IWSLT'14 De-En dataset and the WSJ training dataset. We train an IWSLT'14 language model to compare the parsing performance of the language models with that of the translation models trained on the IWSLT'14 De-En dataset. Additionally, we use WSJ, a staple in parsing and language modeling work, to compare the parsing performance of WSJ-trained language models and IWSLT'14-trained translation models and language models. For WSJ, we use the preprocessing method by \citet{Mikolov2011SUBWORDLM}, which lower-cases the sentences, removes punctuation and replaces the numbers with \textit{N}. For IWSLT, we follow the preprocessing method by \citet{Cettolo2015ReportOT} which lower-cases the sentences and removes XML tags but maintains punctuation and numbers as they are. 

To handle the out-of-vocabulary (OOV) problem, NMT models often use the byte pair encoding compression algorithm (BPE) to form a vocabulary and to tokenize text. The algorithm splits words into sequences of frequent subword units \citep{Sennrich2016NeuralMT}; for example, \textit{green-light} is split into 3 subword units, \textit{gre$^{@@}$}, \textit{en-$^{@@}$}, and \textit{light}. ``\textit{$^{@@}$}" indicates that the next token is continuation of current word.
We train the machine translation models on the IWSLT'14 De-En and IWSLT'17 Zh-En datasets.\footnote{
\url{http://workshop2014.iwslt.org/}
}\footnote{\url{http://workshop2017.iwslt.org/}
} 
We follow the same preprocessing method of \citet{Cettolo2015ReportOT} used in our IWSLT'14 language modeling data. 

We train each model with both a word-based decoder and a BPE-based decoder, which have the same model architecture and hyperparameters, except vocabulary. Since word-based models are poorer at handling OOV tokens, the BLEU scores of the word-based models are expected to be lower than that of BPE-based models. However, it is necessary to train the word-based model for parsing comparison, as the reference parses are word-based. For the De-En dataset, we train BPE-to-BPE and word-to-word translation models. For the Zh-En dataset, since Chinese is harder to accurately tokenize at the word level, we train BPE-to-BPE and BPE-to-word models.  We do not use BPE in language modeling experiments. 

For BPE-based models, we use a BPE vocabulary size of 10,000. We apply BPE to the preprocessed datasets to create additional datasets with subwords. To calculate the F1 for BPE-based decoder, we apply BPE to the reference parses produced by Stanford parser.

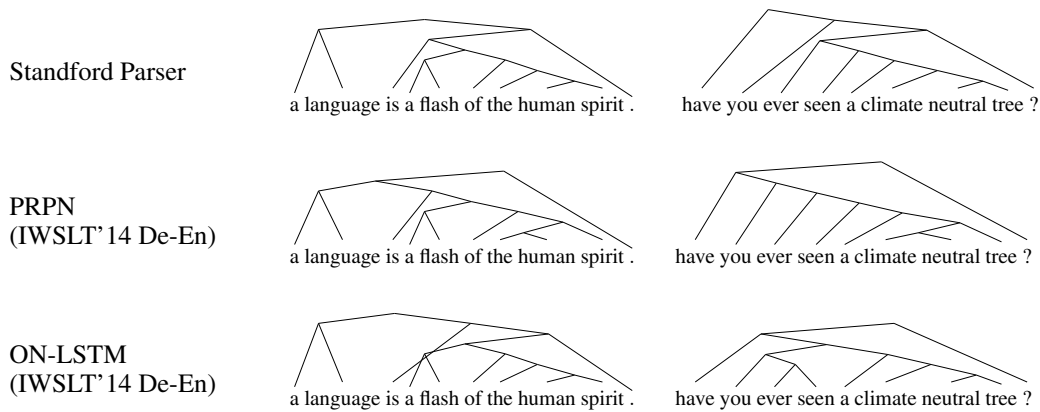
\begin{figure*}[!t]

	\centering
	
	\hfill
	\begin{minipage}{0.2\textwidth}
	Standford Parser
	\end{minipage}
	\begin{minipage}{0.78\textwidth}
	\scalebox{0.8}{
	\begin{forest}
		shape=coordinate,
		where n children=0{
			tier=word
		}{},
		nice empty nodes
		[  [  [ a ]  [ language ]  ]  [  [  [ is ]  [  [  [ a ]  [ flash ]  ]  [  [ of ]  [  [ the ]  [  [ human ]  [ spirit ]  ]  ]  ]  ]  ]  [ $.$ ]  ]  ] 
	\end{forest}}
	\hspace{1.0em}
	\scalebox{0.8}{
	\begin{forest}
		shape=coordinate,
		where n children=0{
			tier=word
		}{},
		nice empty nodes
	[  [ have ]  [  [ you ]  [  [  [ ever ]  [  [ seen ]  [  [ a ]  [  [ climate ]  [  [ neutral ]  [ tree ]  ]  ]  ]  ]  ]  [ ? ]  ]  ]  ] 
\end{forest}}
	\vspace{0.3em}
	\end{minipage}
	\hfill
	\\

	\vspace{1.0em}
	
	\hfill
	\begin{minipage}{0.2\textwidth}
	PRPN 
	
	(IWSLT'14 De-En)
	\end{minipage}
	\begin{minipage}{0.78\textwidth}
	\scalebox{0.8}{
	\begin{forest}
		shape=coordinate,
		where n children=0{
			tier=word
		}{},
		nice empty nodes
        [  [  [  [ a ]  [ language ]  ]  [  [ is ]  [  [  [ a ]  [ flash ]  ]  [  [ of ]  [  [  [ the ]  [ human ]  ]  [ spirit ]  ]  ]  ]  ]  ]  [ $.$ ]  ] 
	\end{forest}}\hspace{1.0em}
	\scalebox{0.8}{
	\begin{forest}
		shape=coordinate,
		where n children=0{
			tier=word
		}{},
		nice empty nodes
		[  [  [ have ]  [  [ you ]  [  [ ever ]  [  [ seen ]  [  [ a ]  [  [  [ climate ]  [ neutral ]  ]  [ tree ]  ]  ]  ]  ]  ]  ]  [ ? ]  ]
\end{forest}}
	\vspace{0.3em}
	\end{minipage}
	\hfill
	\\
	
	\vspace{1.0em}
	
	\hfill
    \begin{minipage}{0.2\textwidth}
	ON-LSTM 
	
	(IWSLT'14 De-En)
	\end{minipage}
	\begin{minipage}{0.78\textwidth}
		\scalebox{0.8}{
	\begin{forest}
		shape=coordinate,
		where n children=0{
			tier=word
		}{},
		nice empty nodes
        [  [  [ a ]  [ language ]  ]  [  [ is ]  [  [  [  [ a ]  [ flash ]  ]  [  [ of ]  [  [ the ]  [  [ human ]  [ spirit ]  ]  ]  ]  ]  [ $.$ ]  ]  ]  ] 
	\end{forest}}\hspace{1.0em}
	\scalebox{0.8}{
	\begin{forest}
		shape=coordinate,
		where n children=0{
			tier=word
		}{},
		nice empty nodes
		[  [  [ have ]  [  [  [ you ]  [  [ ever ]  [ seen ]  ]  ]  [  [ a ]  [  [ climate ]  [  [ neutral ]  [ tree ]  ]  ]  ]  ]  ]  [ ? ]  ] 
\end{forest}}
	\vspace{0.4em}
	\end{minipage}
	\hfill
    \\

\caption{\label{fig:trees} \textit{Top}: Parses from Stanford parser. \textit{Middle}: Parses from the word-based PRPN decoder trained on IWSLT'14 De-En. \textit{Bottom}: Parses from the word-based ON-LSTM decoder (layer 3) trained on IWSLT'14 De-En. We observe that the trees produced by both PRPN and ON-LSTM decoders lean slightly towards right branching.
}
\end{figure*}

\subsection{Implementation}

We implement both models within the Fairseq framework.\footnote{
\url{https://github.com/pytorch/fairseq}
} 
We find that these models are very sensitive to the choice of optimizer and learning rate. We initially experiment with different ranges of learning rates (for example, $10^K$ where $K$ ranges in $[-4,1]$), and choose the best set of hyperparameters on the IWSLT'14 De-En development set. After this, we fix the hyperparameters and train each model with five different random seeds. We will release our code upon acceptance.

For language modeling experiments, we use the Adam optimizer with an initial learning rate of 0.001 for PRPN, and the SGD optimizer with an initial learning rate of 0.7 for ON-LSTM. For all machine translation experiments, we use the Adam optimizer with an initial learning rate of 0.0005 for training. We use a beam-size of 5 with length penalty 1 during inference. We use the Compare-MT framework~\citep{akabe14coling} for the sentence length analysis of our machine translation models.
We use version 3.9.2 of the Stanford parser \citep{manning-EtAl:2014:P14-5} to parse the English target sentences of the IWSLT datasets.
For the BPE-based models, we parse the sentences first and then convert the word-based parsed sentences to BPE tokens; we group the BPE tokens within each word into left branching sub-trees. We measure translation quality using BLEU  \citep{Papineni2001BleuAM} and parsing performance using unlabeled constituency F1 score with respect to the Stanford parser.

\subsection{Reference Parses }

Since the set of translated sentences produced by each model is different, we cannot strictly compare the parsing performance of different models using the parses of translated sentences. Additionally, as the translated sentences produced by the model may not be similar to the gold target translation, it is impossible to compare the parsing performance of NMT decoders with corresponding language models using the parses of translated sentences. Therefore, we feed the gold target sentences from the test set into the trained NMT models to get a set of parses of the gold target sentences. We calculate the sentence level F1 scores of the test sets and report the average \textit{F1 (Target)}.

\section{Results}

Table~\ref{tab:result-table} summarizes the performance of each NMT model trained on the IWSLT'14 De-En dataset. The table also reports the parsing F1 and perplexity of each model trained with a language modeling objective on the target (En) side of the IWSLT'14 De-En dataset. 
Table  \ref{tab:result-table} shows that trees induced by the PRPN decoder and the best parsing layer of the ON-LSTM decoder achieve higher median and maximum F1 (Target) than the trees induced by the corresponding language model. This suggests that MT is a better task than language modeling for learning latent constituency trees. Furthermore, as the MT decoders receive additional, potentially helpful signal from the MT encoder during inference, we investigate whether the MT decoders will still perform as competitively as LM when we reduce the additional encoder signal.
To test this, we try to reduce the possible additional signals from the encoder by using a single \texttt{[EOS]} token as the source sentence and feed the sentences from the WSJ test set to the decoders of pre-trained NMT models. The NMT decoders still achieve better or comparable parsing F1 as the corresponding language models even with just \texttt{[EOS]} as the source sentences (Table~\ref{tab:lm-table}). This result suggests 
that machine translation objective generally provides a richer signal for latent tree learning than language modeling, at least for the models that we experiment with.

Turning towards the translation results (Tables \ref{tab:result-table} and \ref{tab:result-table-zh}), for all BPE-based decoders, the variance in BLEU is low ($\leq$0.3 on both De-En and Zh-En) across restarts, except for one LSTM trained on IWSLT'14 De-En because of one outlier case. However, the variance across restarts is slightly higher with word-based decoders: This might be because BPE-based models are better at handling unknown tokens than word-based models. We observe that the latent-tree based decoders are competitive with the LSTM baseline for both BPE and word-based models. PRPN achieves consistently better BLEU than LSTM and ON-LSTM on De-En translation for both word-based and BPE-based models. Although ON-LSTM performs slightly worse than PRPN and LSTM in terms of maximum BLEU score, the median BLEU of ON-LSTM across restarts is on par with the median BLEU of LSTM. Overall, the latent-tree based decoders are able to induce grammar while achieving good downstream task performance in machine translation.

\begin{table}[t]
\small
\centering
\setlength{\tabcolsep}{4 pt} 
\begin{center}
\begin{tabular}{lrrrrrr}
\toprule
& \multicolumn{3}{c}{ \bf F1 (Target) w.r.t } & \multicolumn{3}{c}{ \bf Accuracy on } \\
  \bf Model  &   \bf  LB & \bf RB & \bf GT   & \bf ADJP & \bf NP & \bf PP    \\
 \midrule
   \multicolumn{7}{c}{ \bf Language Model: IWSLT'14   }   \\
 \midrule
 PRPN  &  24.9 & 25.7 & 42.0 & 40.8 & 56.0 &  46.3 	 \\
  ON-LSTM   & --  & --  & -- & -- & -- & --   \\
 \hspace{1.5em}Layer 1  & 18.2 & 46.1  & 24.0	& 18.0 & 23.1  & 13.8  \\
 \hspace{1.5em}Layer 2   & 19.4	& 44.6 & 38.1	&  22.3 & 54.6  &  37.2  \\
 \hspace{1.5em}Layer 3  & 30.7 & 15.7 & 19.9  & 21.0 & 55.3  &  13.2 \\
 \midrule
   \multicolumn{7}{c}{ \bf Translation: IWSLT'14 De-En  }   \\
 \midrule
 PRPN  &  16.8 & 18.5 & \textbf{53.0} & 37.0 & 62.8 & \bf 64.8 	 \\
  ON-LSTM   & --  & --  & -- & -- & -- & --    \\
 \hspace{1.5em}Layer 1  & 11.4 & 45.4  & 42.4	& 35.6 & 51.6  &  24.7  \\
 \hspace{1.5em}Layer 2   & 23.8	& 13.6 & 49.4	& \bf 50.0 & \bf 70.5  &  52.6  \\
 \hspace{1.5em}Layer 3  & 24.4 & 39.8 & 15.4  & 16.4 & 9.6  &  2.7 \\
 \midrule
   \multicolumn{7}{c}{ \bf Translation: IWSLT'17 Zh-En }   \\
 \midrule
 PRPN  & 16.3  &	44.3 &  37.9 & 24.1 & 49.4  &  41.2  \\
 ON-LSTM   & --  & --  & -- & -- & -- & --   \\
 \hspace{1.5em}Layer 1  & 10.6	& 49.5 & 36.8  & 31.1 & 49.5  &  26.9 	 \\
 \hspace{1.5em}Layer 2  & 26.7	& 10.6 & \textbf{47.0} & \bf 38.8 & \bf 62.8  &  \bf 66.4  \\
 \hspace{1.5em}Layer 3 & 18.9 & 41.2 & 15.0  & 31.4 & 49.5  &  26.9  \\
 \bottomrule  
\end{tabular}
\end{center}
\caption{\label{tab:lbrb-F1} F1 score with respect to trivial left-branching (LB), right-branching (RB) baselines, and the ground truth parses from Stanford parser (GT) on the corresponding IWSLT test set. We also report accuracy in identifying constituents labeled as ADJP, NP, or PP by the parser. We choose the word-based PRPN model and the word-based ON-LSTM model with the \textit{median} F1 for each language pair. Bold marks the best performance of each column on the corresponding dataset.} 
\end{table}

\section{Analysis and Discussion}

Our experiments show that MT decoders perform slightly better in parsing than LM even when the encoder input is unavailable, suggesting that they benefit from the additional, potentially more semantically-informed signal during training.
Focusing first on NMT, sentence length analysis using the average sentence-level smoothed BLEU scores (Figure \ref{fig:bleu-length}) shows that PRPN competes with LSTM across all sentence lengths, and ON-LSTM performs better than LSTM for shorter sentences. However, ON-LSTM performs worse than LSTM for longer sentences with length greater than 40. As ON-LSTM uses the consecutive fractions of hidden state neurons to represent the tree nodes, there is an upper bound on the depth of the tree. This could affect its ability to properly track extremely long dependencies. 

In terms of parsing, we find that the variance in F1 across restarts of the LTL models trained with translation objective is much higher than the LTL models that are trained with the language modeling objective (Table \ref{tab:result-table}). The high F1 variance across restarts of NMT's induced parses is also observed in the models trained on the Zh-En dataset (Table \ref{tab:result-table-zh}), indicating that this is not language-pair specific. Therefore, although machine translation is generally a good objective for inducing latent trees, final translation quality on a held-out set is not a good indicator of induced trees' quality.


According to our qualitative analysis, PRPN and ON-LSTM trained with NMT objective are good at discovering noun phrases and prepositional phrases (Table~\ref{tab:lbrb-F1}). For example, we can observe in Figure  \ref{fig:trees} that both PRPN and ON-LSTM correctly identify noun phrases such as ``\textit{a language}", ``\textit{a flash of the human spirit}". We also notice that the Stanford parser wrongly group "neutral" and "tree" first in ``\textit{a climate neutral tree}", while PRPN correctly groups "climate neutral" correctly. This also indicates using Stanford parser as the standard parse might not be the best way to evalute the parsing performance; therefore, we also perform further evaluation on parsing using expert-annotated WSJ test set (Table \ref{tab:lm-table}). We also find that both PRPN and ON-LSTM decoders tend to group the BPE subwords under the same constituents in many cases. We observe that PRPN correctly group ``\textit{chemo$^{@@}$ syn$^{@@}$ thesis}", ``\textit{gre$^{@@}$ en-$^{@@}$ light}", etc. (Figure~\ref{fig:bpetrees} in the appendix). Although ON-LSTM fails to correctly group the BPE subwords into words in these examples, the subwords are still put under the same intermediate level constituent such as ``( \textit{the} ( \textit{ris$^{@@}$} (\textit{ing billion} ) ) )". 

Additionally, in Table  \ref{tab:result-table}, we observe that, in our language modeling experiments, layer 2 of ON-LSTM has consistently higher F1 score than layers 1 and 3 for all 5 random seeds which agrees with the findings of \citet{shen2018ordered}. However, in our machine translation experiments, the best parsing layer of ON-LSTM varies across  restarts, though layer 1 has highest median and maximum F1.

We also observe that the relative difference in F1 between layers 2 and 3 of word-based ON-LSTM is much higher than that of BPE-based ON-LSTM on both De-En and Zh-En datasets (Table  \ref{tab:result-table} and  \ref{tab:result-table-zh}). 
Since layer 3 of the BPE-based models might need to learn syntax information to correctly predict the subwords, we hypothesize that this alleviates the need for layer 2 to learn syntax in the BPE-based models. Thus, the relative parsing performance of layer 2 of the BPE-based ON-LSTM decoders is lower than that of word-based ON-LSTM decoders. This also suggests that syntax information is learned jointly by all layers of ON-LSTM, and that  layer-by-layer F1 score of ON-LSTM may not be a good indication of the level of syntactic information learned by the entire model.

\section{Conclusion}
\label{sec:conclusion}

We train two high-performing LTL models---PRPN and ON-LSTM---that are designed for language modeling, as the decoders of neural machine translation models. Our experiments show that these latent tree-based neural machine translation decoders successfully learn to induce reasonably linguistically plausible constituency trees while achieving good downstream task performance on translation. Furthermore, our experiments demonstrate that PRPN and ON-LSTM trained as machine translation decoders perform competitively or better than respective language models for latent tree learning. We therefore conclude that machine translation is a more signal-rich task for inducing trees than language modeling. However, the high variance in parsing F1 of MT decoders is alarming: does it happen only in the models we investigated, or is it generally true for latent tree learning with translation objective? What kind of consistent linguistic structures these models learn? A more thorough analysis of the models and the parses produced would be needed to understand these open questions in future work.  
\section*{Acknowledgments}

This work was supported by Samsung Research under the project \textit{Improving Deep Learning using Latent Structure} and from the donation of a Titan V GPU by NVIDIA Corporation.

\bibliography{acl2019}
\bibliographystyle{aaai}

\end{document}